%% file: main.tex
\documentclass{article}

\usepackage{iftex}
\ifPDFTeX
  \usepackage{cmap}
  \usepackage[T1]{fontenc}
  \usepackage{firmwm_arxiv,times}
\else
  \usepackage{firmwm_arxiv}
  \usepackage{fontspec}
\fi
\input{math_commands.tex}

\usepackage{amsmath,amssymb,mathtools}
\renewcommand{\eqref}[1]{(\ref{#1})}
\usepackage{booktabs}
\usepackage{graphicx}
\usepackage{algorithm}
\usepackage{algpseudocode}
\usepackage{microtype}
\usepackage{xcolor}
\usepackage{multirow}
\usepackage{placeins}
\usepackage{tikz}
\usetikzlibrary{arrows.meta,positioning,fit,backgrounds}
\usepackage{hyperref}
\hypersetup{hidelinks}
\usepackage{url}

\input{numbers.tex}

\definecolor{firmblue}{RGB}{50,101,168}
\definecolor{firmgreen}{RGB}{55,126,85}
\definecolor{firmorange}{RGB}{202,112,38}
\definecolor{lightblue}{RGB}{232,241,250}
\definecolor{lightgreen}{RGB}{233,245,237}
\definecolor{lightorange}{RGB}{251,240,226}

\newcommand{\method}{FIRM-WM}
\newcommand{\DF}{\mathcal{D}_{\mathrm{F}}}
\newcommand{\DI}{\mathcal{D}_{\mathrm{I}}}
\newcommand{\tightparagraph}[1]{\par\smallskip\noindent\textbf{#1}\enspace}

\title{\upshape FIRM-WM: State-factorized\\
factual--interventional recurrent modeling\\
for reward-free visual planning}

\author{%
  {\normalsize
  Yilun Wu\textsuperscript{1,2}\quad
  Yunjian Zhang\textsuperscript{2}\quad
  Aobo Li\textsuperscript{2}\\[3pt]
  Mujiangshan Wang\textsuperscript{3,2,*}\quad
  Haitao Wu\textsuperscript{1}\quad
  Aqiang Zhang\textsuperscript{4,5}}\\[8pt]
  \textsuperscript{1}Saint Petersburg State University, Saint Petersburg, Russia\\
  \textsuperscript{2}Shenzhen Kaihong Digital Industry Development Co., Ltd., Shenzhen, China\\
  \textsuperscript{3}Chinese Academy of Sciences (CAS) -- Shenzhen Institute of Advanced Technology\\
  \textsuperscript{4}Harbin Institute of Technology, Harbin, China\\
  \textsuperscript{5}Chongqing Research Institute of HIT, Chongqing, China\\[6pt]
  \texttt{wuyilun310@gmail.com}\quad\texttt{yunjzhang@xmu.edu.cn}\\
  \texttt{ali09@qub.ac.uk}\quad\texttt{wuhaitao2023@gmail.com}\\[3pt]
  \textsuperscript{*}Corresponding author: Mujiangshan Wang (\texttt{mjs.wang@siat.ac.cn})%
}
\date{}
\hypersetup{%
  pdftitle={FIRM-WM: State-factorized factual--interventional recurrent modeling for reward-free visual planning},
  pdfauthor={Yilun Wu; Yunjian Zhang; Aobo Li; Mujiangshan Wang; Haitao Wu; Aqiang Zhang}%
}

\begin{document}
\maketitle

\begin{abstract}
Reward-free latent world models can learn from offline videos and solve new image-goal tasks by optimizing actions through predicted latent futures.  This setting places two demands on the planning state: its coordinates must be comparable with a goal image.  Moreover, its dynamics must retain velocity, motion trend, contact, and other history-dependent information beyond those goal coordinates.  Offline training creates a second mismatch: each recorded trajectory reveals one factual future, whereas a sampling-based planner compares many actions that were not taken from the same state.  We introduce \method{} (\emph{Factual--Interventional Recurrent World Model}), a compact pixel world model designed around these two gaps.  Its recurrent state separates a typed, goal-comparable configuration from a 128-dimensional dynamic fiber used for prediction but excluded from the terminal goal cost.  Broad factual trajectories provide state coverage, while common-reset intervention branches provide observed outcomes for alternative action sequences.  Before executing each branch, we reset the environment and restore the same recorded values exposed by the environment's state-setting interface.  Under matched CEM planning and three independent full-pipeline seeds, \method{} reaches \TwoRoomFIRM\% on TwoRoom, \ReacherFIRM\% on Reacher, and \CubeFIRM\% on OGBench-Cube, compared with \TwoRoomLeWM\%, \ReacherLeWM\%, and \CubeLeWM\% for LeWM.  The deployed model uses \FIRMParamsMin--\FIRMParamsMax\,M parameters and records 2.13--11.60$\times$ lower planning time on these tasks.
\end{abstract}

\section{Introduction}

World models promise an appealing separation between learning and acting: learn task-agnostic dynamics from offline experience, then compose those dynamics with a goal at deployment.  Recent latent predictive models make this practical from pixels by replacing image reconstruction with prediction in a compact representation space \citep{hafner2019planet,zhou2024dinowm,sobal2025rewardfree,maes2026lewm}.  Given a current observation and a goal image, these systems roll candidate action sequences through a learned predictor and optimize a latent terminal cost with the cross-entropy method (CEM) \citep{rubinstein1999cem}.  They require no task reward during representation learning and plan directly with the learned world model at deployment.

Image-goal planning first creates a representation tension.  The planner needs a compact coordinate system in which a predicted future can be compared directly with a static goal image.  A typed task configuration---such as position or joint angle---provides such a geometry, but is generally not a sufficient dynamical state: two histories can share the same configuration while differing in velocity, direction of motion, contact, or future controllability.  Conversely, using an unrestricted visual latent for both prediction and goal comparison mixes these dynamical distinctions with appearance variation that need not affect the terminal goal cost.  A useful planning state must therefore be goal-comparable without discarding the history required to predict its evolution.

Offline learning creates a second, complementary gap.  A recorded trajectory contains the future produced by the action that was actually executed at each history; it does not show what the same system state would have done under other actions.  CEM, by contrast, evaluates hundreds of alternative sequences from the current history.  Factual training therefore constrains the transition at the executed action but leaves the outcomes of alternative actions underdetermined.  This gap is distinct from long-horizon error accumulation: limited action variation at a fixed state can leave controlled dynamics unidentified from the recorded data \citep{zhang2026controlledidentifiability}.

\method{} addresses the two gaps with a factual--interventional recurrent state factorization.  It represents the planning state as
\begin{equation}
    s_t=(c_t,m_t),
    \qquad c_t\in\mathcal{C},\quad m_t\in\mathbb{R}^{128},
    \label{eq:state}
\end{equation}
where $c_t$ is a typed physical configuration that can be compared with a static goal, and $m_t$ is a learned dynamic fiber that retains the history-dependent information needed to predict future configurations.  The transition updates both variables recursively, but the terminal goal cost compares only $c_t$.  Configuration and fiber therefore serve distinct roles: the former is the variable used by the terminal goal cost, while the latter completes the predictive state.

The model is trained from two complementary sources.  Broad factual sequences cover the state distribution.  For each intervention anchor, we reset the environment and restore the same recorded values of all variables exposed through the environment's state-setting interface before executing each alternative action sequence.  We refer to the resulting trajectories as \emph{common-reset intervention branches}.  ``Common reset'' denotes equality of the restored interface variables across branches, not identity of the complete simulator state or internal solver memory.  These trajectories directly supervise how alternative actions change the configuration and auxiliary dynamics.  The distinction is important: related physical-grounding methods decode state from latents \citep{yan2026psgjepa} or separate predictions under perturbed actions \citep{zeng2026phylatent}, whereas our intervention targets are observed environment trajectories produced by actions executed from common resets.  Figure~\ref{fig:method} shows the resulting image-goal planning architecture, and Table~\ref{tab:landscape} positions its supervision, planner, and active model size against representative pixel world models.

Our contributions are:
\begin{itemize}
    \item We formulate the \emph{factual--interventional gap} in reward-free latent planning: factual prediction constrains the recorded action, while CEM depends on alternative-action outcomes from the current state.
    \item We introduce a compact recurrent state factorization that separates a typed goal configuration from a learned dynamic fiber, preserving history-dependent dynamics while making candidate rollouts and terminal scores interpretable in task coordinates.
    \item We train one recurrent transition with broad factual trajectories and common-reset intervention branches, using a shared multi-horizon physical loss and no reward, success, ranking, or planner-generated label.
\end{itemize}

\section{Related Work}

\paragraph{Latent world models for planning.}
PlaNet and subsequent model-based RL agents learn latent dynamics from pixels, but use reward prediction and task-specific behavior learning \citep{hafner2019planet,hafner2023dreamerv3,hansen2024tdmpc2}.  Reward-free visual planners instead learn from offline trajectories and optimize new image goals at deployment.  DINO-WM predicts frozen DINOv2 patch features \citep{oquab2023dinov2,zhou2024dinowm}; PLDM studies when latent planning is preferable to offline goal-conditioned RL \citep{sobal2025rewardfree}; and LeWM trains an end-to-end JEPA from pixels with a compact predictor and isotropic latent regularization \citep{assran2023ijepa,maes2026lewm}.  Fast-LeWM predicts action prefixes in parallel to reduce autoregressive rollout cost \citep{gao2026fastlewm}.  These approaches typically ask one latent representation to support both action-conditioned prediction and terminal goal comparison; LeWM, for example, ranks candidates by squared distance between the predicted terminal embedding and the encoded goal \citep{maes2026lewm}.  FIRM-WM instead uses a typed configuration for terminal goal comparison and a separate history-dependent fiber for prediction.  Candidate rollouts and terminal scores are therefore directly interpretable in task coordinates rather than only through an unrestricted embedding distance.

\paragraph{Physical structure and controlled identifiability.}
Physical latent models have long sought coordinates that simplify control \citep{jaques2021newtonianvae}.  Recent work adds physical probes or auxiliary objectives to JEPA world models.  PSG-JEPA grounds individual latents and latent changes in robot state \citep{yan2026psgjepa}.  Concurrent PhyLatent targets invariance, physical identifiability, and predicted branch separation while removing its auxiliary heads at deployment \citep{zeng2026phylatent}.  These methods establish that global non-collapse does not imply control-relevant local geometry.  Recent theory casts Gaussian anti-collapse as a distributional anchor whose residual orthogonal symmetry matches Euclidean planning, while warning that exact isotropy in an overcomplete latent can crumple the data manifold \citep{zhang2026latentgeometry}.  Physical coordinates provide an interpretable goal geometry, but using them as the entire state can discard the dynamic context required for closed prediction.  FIRM-WM therefore pairs a typed, goal-comparable configuration with an unrestricted dynamic fiber: the former defines the terminal cost, while the latter distinguishes histories with the same configuration but different futures.  Weak conditional action excitation leaves those futures underdetermined for alternative actions \citep{zhang2026controlledidentifiability}; executed common-reset interventions supervise observed outcomes under sampled alternative actions.

\paragraph{Planning in learned latent spaces.}
Latent MPC can be improved through temporal hierarchy \citep{zhang2026hierarchical}, learned subgoal proposals \citep{cheng2026sage}, amortized inverse dynamics \citep{nguyen2026gcidm}, or cost-ranking objectives \citep{chahe2026monotone}.  These methods improve how candidate trajectories are proposed or ranked.  FIRM-WM instead changes the state and transition supplied to the search: CEM compares predicted goal configurations, while the dynamic fiber carries information needed to produce those predictions under alternative actions.  Our comparisons hold the CEM procedure and candidate budget fixed.

\begin{table*}[t]
\centering
\small
\setlength{\tabcolsep}{5pt}
\caption{Representative pixel world models.}
\label{tab:landscape}
\resizebox{\textwidth}{!}{%
\begin{tabular}{lcccll}
\toprule
Method & Pixel deployment & Reward-free WM & Common-reset branches & Planning & Active parameters \\
\midrule
RC-aux \citep{li2026rcaux} & \checkmark & \checkmark & -- & reachability-aware CEM & $\sim$15.7\,M$^\dagger$ \\
SAGE \citep{cheng2026sage} & \checkmark & \checkmark & -- & subgoal prior + CEM & $\sim$49.3\,M$^\dagger$ \\
LeWM \citep{maes2026lewm} & \checkmark & \checkmark & -- & CEM & $\sim$\LeWMPaperParams\,M \\
Fast-LeWM \citep{gao2026fastlewm} & \checkmark & \checkmark & -- & CEM & $\sim$18\,M \\
PSG-JEPA \citep{yan2026psgjepa} & \checkmark & \checkmark & -- & CEM / policy & NR \\
PhyLatent \citep{zeng2026phylatent} & \checkmark & \checkmark & -- & CEM & LeWM backbone \\
\textbf{FIRM-WM (ours)} & \checkmark & \checkmark & \checkmark & CEM & \FIRMParamsMin--\FIRMParamsMax\,M \\
\bottomrule
\end{tabular}%
}
\vspace{2pt}
\parbox{0.96\textwidth}{\footnotesize NR: not reported in a directly comparable active-deployment accounting. $\dagger$ Active-stack estimates include the reported $\sim$15M LeWM backbone plus the online RC-aux head (0.675M) or SAGE subgoal and action generators (34.31M).}
\end{table*}

\section{Problem Setting: Reward-Free Image-Goal Planning}

We study visual control from offline trajectories without reward supervision.  At deployment, the controller receives a goal image $o^g$ and the causal history
\begin{equation*}
    h_t=(o_{t-L+1:t},a_{t-L+1:t-1}),
\end{equation*}
which contains the last $L$ images and the raw actions executed between them.  The symbol $h_t$ always denotes the currently available history: during training it is taken from a recorded trajectory, whereas during deployment it is formed from live observations and executed actions.  To reach the goal, CEM evaluates a horizon-$H$ sequence of macro actions $U=(u_0,\ldots,u_{H-1})$ by rolling a learned world model forward, then executes the first macro action and replans from the next observation.

The central difficulty is a mismatch between what the offline data record and what the planner asks the model to predict.  The factual dataset $\DF$ contains image--action trajectories collected before training.  At a history sampled from one of those trajectories, it provides the future following the action sequence that was actually recorded; we call this the \emph{factual} continuation.  During deployment, CEM instead proposes many candidate sequences from the current live history, predicts their futures with the world model, and repeatedly updates its candidates toward lower predicted goal cost.  Most of these candidates have no corresponding real outcome in $\DF$ from exactly the same physical state.  We call their predicted futures \emph{counterfactual planner queries}.  The recorded continuation therefore does not by itself determine the outcomes of the alternatives that CEM compares when choosing an action.

A world model for this setting must therefore satisfy three properties.  Its predicted terminal state must contain a coordinate that can be compared directly with the goal image; its recurrent state must retain the history-dependent information needed to predict how that coordinate evolves; and its training signal must constrain outcomes under actions beyond the single factual continuation observed at each history.  Section~\ref{sec:method} instantiates these requirements with a configuration--fiber recurrent state and complementary factual and common-reset intervention rollouts.

\section{FIRM-WM: Pixel-Based Counterfactual World Modeling}
\label{sec:method}

\begin{figure*}[t]
\centering
\includegraphics[width=0.99\textwidth]{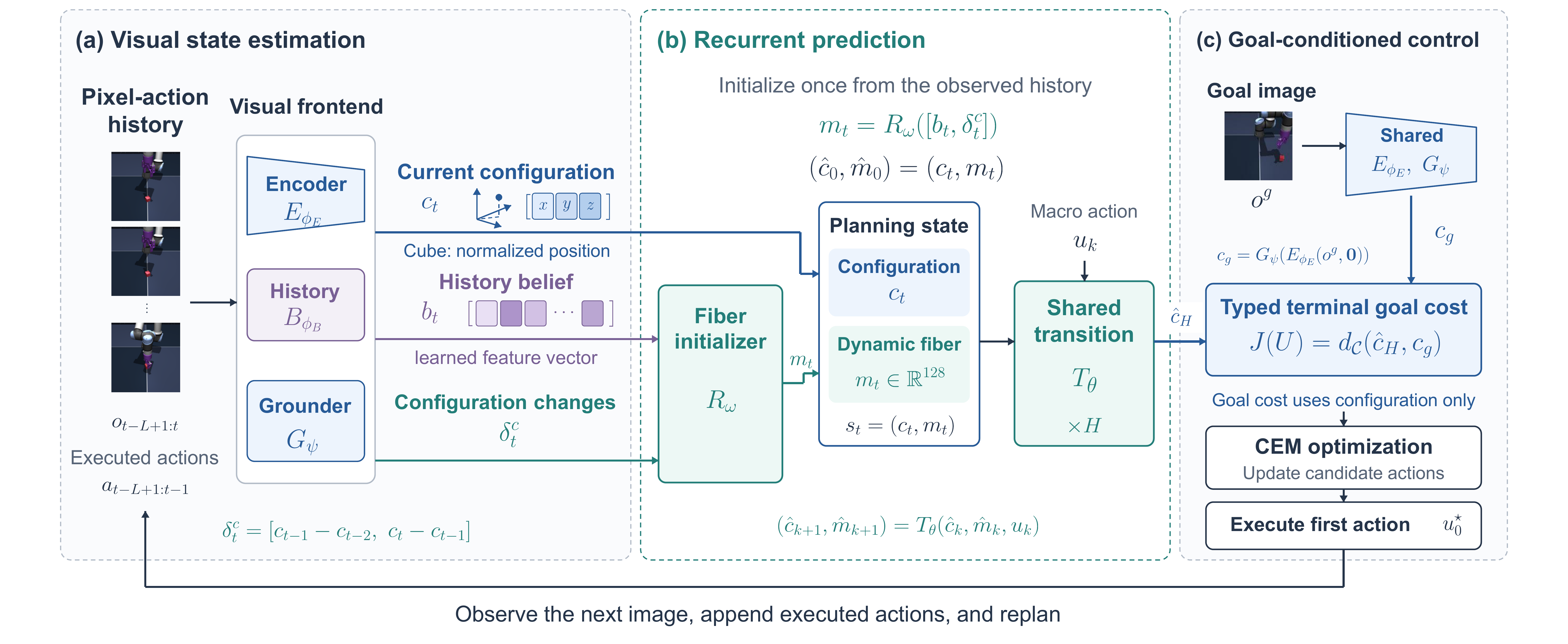}
\caption{\textbf{FIRM-WM image-goal planning architecture.}  The pixel--action history is mapped by the encoder, causal history module, and grounder to the current configuration $c_t$ and belief $b_t$.  The history-to-fiber initializer constructs $m_t$, giving the recurrent state $s_t=(c_t,m_t)$.  For each candidate macro-action sequence, the single shared transition recursively updates configuration and fiber for $H$ steps.  The shared encoder and grounder map the goal image to $c_g$, and the typed terminal distance $d_{\mathcal C}(\hat c_H,c_g)$ supplies the CEM objective $J(U)$.  The controller executes the first optimized macro action, appends the next observation and executed action to the causal history, and replans.  The fiber affects prediction but is excluded from the goal cost.}
\label{fig:method}
\end{figure*}

FIRM-WM realizes these requirements in one recurrent planning pipeline.  Spatial tokens and a causal belief infer the current situation from pixels; a typed configuration is compared with the goal image by the terminal cost; and a compact dynamic fiber retains the additional history needed for prediction.  The shared transition learns from broad factual rollouts together with common-reset intervention branches, then supplies counterfactual configuration rollouts to CEM.  Figure~\ref{fig:method} summarizes this image-goal planning architecture.

\subsection{Pixel tokens and causal history belief}

A current image reveals scene geometry but not, by itself, how that scene is evolving.  We preserve its spatial organization with a small set of local visual tokens,
\begin{equation}
    p_t=E_{\phi_E}(o_t,\Delta o_t),
    \qquad \Delta o_t=o_t-o_{t-1}.
\end{equation}
Here, $E_{\phi_E}$ denotes the pixel frontend: a spatial tokenizer for TwoRoom, Push-T, and Cube, and a compact geometry encoder for Reacher.  Its output encodes where task-relevant structures appear in the current frame.  To recover motion information that is ambiguous in a single image, the causal history attention module fuses the recent token sequence with the actions executed between frames,
\begin{equation}
    b_t=B_{\phi_B}(p_{t-L+1:t},a_{t-L+1:t-1}).
\end{equation}
Thus $p_t$ describes the current visual arrangement, whereas $b_t$ recovers motion evidence unavailable from one frame and summarizes the visual--action history needed to predict its evolution.  The task-typed frontend architectures, token dimensions, and history lengths are reported in Appendix~\ref{app:interfaces}; all tasks expose the same token--belief interface to the recurrent state below.

\subsection{Configuration--fiber recurrent dynamics}

The goal-comparable component of the recurrent state is a typed configuration inferred from a canonical encoding of the current image,
\begin{equation}
    c_t=G_\psi(E_{\phi_E}(o_t,\mathbf 0))\in\mathcal{C}.
\end{equation}
For the spatial encoders, the zero second input removes the frame-difference channel from this configuration readout.  Depending on the task, $c_t$ contains position, periodic joint angles, or object pose components; Appendix~\ref{app:interfaces} specifies each interface.  The grounder $G_\psi$ is supervised by physical configuration labels during training and shared between current and goal images.  Its output therefore supplies the typed geometry in which terminal predictions are compared with the goal.

We train the pixel frontend and configuration grounder before the reported recurrent-dynamics refinement.  The pixel frontend uses task-typed visual and configuration-preserving objectives, while the grounder regresses the physical configuration coordinates defined in Appendix~\ref{app:interfaces}.  The causal history module is then either frozen or jointly refined with the recurrent transition according to the task schedule.  The history-to-fiber initializer and unified transition are trained recursively on factual and common-reset intervention rollouts.  Appendix~\ref{app:disclosure} reports each module's initialization, objective, optimizer, update count, freezing rule, and checkpoint-selection criterion for every environment.

Configuration alone is not a sufficient dynamical state.  Two histories can have the same $c_t$ while differing in velocity, motion direction, contact mode, or other variables that change their action-conditioned futures.  FIRM-WM retains this residual dynamic context in a compact fiber.  A history-to-fiber encoder constructs the initial fiber state for each candidate rollout from the causal belief and recent changes in configuration,
\begin{equation}
    \delta^c_t=\left[c_{t-1}-c_{t-2},\;c_t-c_{t-1}\right],
    \qquad m_t=R_\omega([b_t,\delta^c_t])\in\mathbb{R}^{128},
    \label{eq:fiberinit}
\end{equation}
where $[\cdot,\cdot]$ denotes feature concatenation.  The encoder $R_\omega$ is evaluated once at the current observed history to initialize $m_t$; subsequent fiber states are produced recursively by the transition.  For an angular coordinate $\vartheta$, $c_t$ stores $(\sin\vartheta,\cos\vartheta)$.  Differences in $\delta^c_t$ are therefore taken between sine--cosine pairs and do not jump when the angle crosses $\pm\pi$.  The resulting planning state $s_t=(c_t,m_t)$ separates two roles: $c_t$ provides goal geometry, while $m_t$ retains history-dependent information that is excluded from the goal cost but needed to predict the next recurrent state.  For one candidate rollout, set $(\hat c_0,\hat m_0)=(c_t,m_t)$.  The recurrent transition updates both components,
\begin{equation}
    (\hat c_{k+1},\hat m_{k+1})
    =T_\theta(\hat c_k,\hat m_k,u_k),
    \qquad k=0,\ldots,H-1,
    \label{eq:closedtransition}
\end{equation}
where $u_k$ is the $k$th macro action in the candidate sequence $U$.  Across all tasks, $T_\theta$ remains one recurrent operator with one configuration head and one fiber head.  TwoRoom, Reacher, and Cube use a compact residual-MLP hidden update.  Push-T retains the same state, action, and output interface but adds an action-conditioned spatial residual inside the same hidden update to preserve contact geometry.  Under our settings, CEM evaluates this operator over $15{,}000$--$45{,}000$ candidate--step pairs per solve, without repeated token-level attention in the planning inner loop.  The transition applies its increments to Euclidean position coordinates, sine--cosine angle coordinates, and the fiber, so both the predicted configuration and dynamic context evolve throughout the rollout.  During training, an auxiliary decoder $D_\rho$ maps $\hat m_k$ to an environment-typed dynamic target, $\hat\chi_k=D_\rho(\hat m_k)$, encouraging the fiber to retain variables such as velocity or contact.  The decoder and target are discarded after training and never enter the goal cost.  Network widths and task-typed residual details are given in Appendix~\ref{app:interfaces}.

\subsection{Learning from factual and interventional rollouts}

Learning the transition requires two complementary kinds of evidence.  Factual rollouts cover the broad range of states present in the offline trajectories, but provide little action variation at any one history.  Common-reset interventions concentrate on fewer histories but reveal how alternative actions change their futures.

A factual record is a rollout $\xi^{\mathrm F}=(h_i,U_i^{\mathrm F},c_{i,0:H},\chi_{i,0:H})$ extracted from an offline trajectory.  Here $c_{i,k}$ is the goal-comparable configuration defined above, while $\chi_{i,k}$ is an environment-typed auxiliary dynamics vector, such as velocity, displacement, contact, or the remaining robot state.  The factual dataset $\DF$ contains these records across the state coverage of the offline trajectories.

To collect common-reset intervention data, we select a time index $i$ from a training episode and save two items: the observation--action history $h_i$ and the recorded reset condition $x_i^{\mathrm{sim}}$.  Here, $x_i^{\mathrm{sim}}$ contains only the physical variables exposed by the environment's state-setting interface.  For each alternative macro-action sequence $U_{ij}$, we first reset the environment, restore $x_i^{\mathrm{sim}}$, execute $U_{ij}$, and record the resulting configuration and auxiliary-dynamics trajectories.  Repeating this procedure for $K_i$ action sequences gives
\begin{equation}
 \mathcal{I}_i=
 \left\{\xi^{\mathrm I}_{ij}
 =\left(h_i,U_{ij},c_{ij,0:H},\chi_{ij,0:H}\right)\right\}_{j=1}^{K_i}.
 \label{eq:intervention}
\end{equation}
These branches begin from the same restored interface variables and differ in their action sequences.  The symbol $\xi^{\mathrm I}_{ij}$ denotes one resulting training record, and $\DI$ is the collection of these records.  Thus $\DF$ provides broad state coverage, while $\DI$ provides observed futures for alternative actions under common reset conditions.  Appendix~\ref{app:disclosure} specifies the reset variables, anchor counts, action sampling, and branch counts for every task; Appendix~\ref{app:interfaces} gives the task-specific contents of $\chi$ and the raw-to-macro timing.  A factual-suffix reset-replay audit over 1,210 eligible training anchors keeps the P95 task-normalized configuration discrepancy below 0.15 success-tolerance units at every macro horizon (Table~\ref{tab:reset-replay}).  The only auxiliary decoder in this stage is $D_\rho$, which maps $m_k$ to $\chi_k$ during training and is absent from the planner.

For environment $e$, recurrent stage $r$ activates sources $\mathcal A_{e,r}\subseteq\{\mathrm F,\mathrm I\}$ and applies $N_{e,r}$ updates to parameters carried from the preceding stage.  Algorithm~\ref{alg:training} and Appendix~\ref{app:disclosure} give the complete module and source schedules.

For either source $\sigma\in\{\mathrm F,\mathrm I\}$, the rollout loss has three functional parts,
\begin{equation}
\ell_\sigma(\mathcal B)
=\underbrace{\ell_{\sigma,\mathrm{cfg}}(\mathcal B)}_{\text{future configuration}}
+\underbrace{\ell_{\sigma,\mathrm{disp}}(\mathcal B)}_{\text{configuration displacement}}
+\underbrace{\ell_{\sigma,\mathrm{dyn}}(\mathcal B)}_{\text{fiber dynamics}}.
\label{eq:loss-groups}
\end{equation}
The configuration term supervises the absolute future $c_{1:H}$ using squared Euclidean error for position coordinates and squared distance between sine--cosine pairs for angles.  The displacement term matches changes between configurations, so the loss measures the effect of the executed action rather than allowing static scene content to dominate.  The dynamics term supervises $D_\rho(\hat m_k)$ with the training-only targets $\chi_k$.  Appendix~\ref{app:supervision} defines the displacement operator $\Lambda_\kappa$ and gives the exact tail reductions, initial-fiber term, and component weights used by each task and data source.  In the final recurrent stage of every environment, both sources update one transition through the joint objective
\begin{equation}
    \mathcal{L}_{\mathrm{FIRM}}
    =\mathbb{E}_{\mathcal B_{\mathrm F}\sim\operatorname{Batch}(\DF)}
       [\ell_{\mathrm F}(\mathcal B_{\mathrm F})]
    +\mathbb{E}_{\mathcal B_{\mathrm I}\sim\operatorname{Batch}(\DI)}
       [\ell_{\mathrm I}(\mathcal B_{\mathrm I})],
    \label{eq:total}
\end{equation}
where $\operatorname{Batch}(\mathcal D)$ denotes a minibatch drawn from dataset $\mathcal D$.  The full-pipeline factual-only and intervention-only variants in Appendix~\ref{app:ablation} restrict every task-training stage to $\DF$ or $\DI$, respectively, while retaining the architecture, initialization scheme, optimization schedule, and paired evaluation protocol.

Equation~\eqref{eq:total} therefore joins state coverage with action coverage at common reset conditions: factual data teach how the system evolves along recorded trajectories, while interventions teach how alternative actions change that evolution after restoring the same interface variables.  Earlier recurrent stages use the active source sets listed in Appendix~\ref{app:disclosure}.  The complete recurrent-transition update count is $N_{\mathrm{rec}}^{(e)}=\sum_{r=1}^{R_e}N_{e,r}$.  The intervention records in Equation~\eqref{eq:intervention} are constructed before recurrent training.  Simulator resets therefore occur only during data collection; Algorithm~\ref{alg:training} operates on completed factual and intervention records.

\begin{algorithm}[t]
\caption{Complete recurrent-transition training across environment-specific stages}
\label{alg:training}
\small
\begin{algorithmic}[1]
\Require For environment $e$: completed data, source sets $\mathcal A_{e,r}$, update counts $N_{e,r}$, fixed input modules, recurrent initializer, AdamW optimizer, and horizon $H$
\Ensure Deployment modules $(E_{\phi_E},B_{\phi_B},G_\psi,R_\omega,T_\theta)$
\State $\Theta_{\mathrm{train}}\coloneqq(\omega,\theta,\rho)$
\For{$r=1,\ldots,R_e$}
    \State Load fixed $(E_{\phi_E},B_{\phi_B},G_\psi)$ and $\{\mathcal D_\sigma:\sigma\in\mathcal A_{e,r}\}$
    \State If needed, add zero-initialized input weights without changing existing weights
    \For{$n=1,\ldots,N_{e,r}$}
        \For{$\sigma\in\mathcal A_{e,r}$}
            \State Sample $\mathcal B_\sigma\sim\operatorname{Batch}(\mathcal D_\sigma)$; infer $(\hat c_0^\sigma,\hat m_0^\sigma)$; roll out $T_\theta$ for $H$ steps
        \EndFor
        \State $\mathcal L\gets\sum_{\sigma\in\mathcal A_{e,r}}\ell_\sigma(\mathcal B_\sigma)$
        \State $\Theta_{\mathrm{train}}\gets\operatorname{AdamWStep}(\Theta_{\mathrm{train}},\nabla_{\Theta_{\mathrm{train}}}\mathcal L)$
    \EndFor
\EndFor
\State Discard $D_\rho$ and return $(E_{\phi_E},B_{\phi_B},G_\psi,R_\omega,T_\theta)$
\end{algorithmic}
\end{algorithm}

\subsection{Reward-free latent MPC}

First, the goal image is encoded by the same pixel encoder and grounder as the current observation,
\begin{equation}
    p^g=E_{\phi_E}(o^g,\mathbf 0),
    \qquad c_g=G_\psi(p^g).
\end{equation}
Second, Equation~\eqref{eq:closedtransition} recursively predicts the terminal configuration $\hat c_H(U)$ for every candidate macro-action sequence $U$.  Third, the planner scores that candidate with the task-typed configuration distance
\begin{equation}
    J(U;h_t,o^g)=d_{\mathcal C}\!\left(\hat c_H(U),c_g\right),
    \label{eq:planning}
\end{equation}
where $d_{\mathcal C}$ respects Euclidean and periodic components of the task configuration.  The fiber affects the predicted future but is not itself compared with the goal.  Unlike a terminal distance over an unrestricted learned embedding, this objective makes candidate ranking directly interpretable: its score is expressed through named quantities such as position and joint-angle error, with the exact decomposition given in Appendix~\ref{app:interfaces}.  Finally, CEM minimizes Equation~\eqref{eq:planning}; the controller executes the first optimized macro action, observes the next image, and replans.

\section{Experiments and Results}

\subsection{Experimental setup}

We evaluate image-goal control on TwoRoom navigation, two-joint Reacher, Push-T planar manipulation, and OGBench-Cube manipulation \citep{park2024ogbench}.  Each environment uses fixed training and validation splits.  The training split supplies factual trajectories and intervention anchors, the validation split supplies held-out representation and rollout metrics, and final paired control is measured on a fixed set of 100 trials.

FIRM-WM and LeWM use the same task-specific CEM protocol and the same 100 paired trials in each environment.  Appendix~\ref{app:interfaces} reports the complete candidate, iteration, elite, horizon, action-block, and extraction settings.  The Cube appendix additionally reports the public seed-42 50-episode reproduction.

\tightparagraph{Training variance.}
To assess training stability, we retrain the complete \method{} pipeline with three independent random seeds.  Each run initializes all trainable stages independently while holding the training split, validation split, intervention data, training recipe, and paired evaluation trials fixed.

\tightparagraph{Baselines and access.}
LeWM is evaluated with its released checkpoint and solver \citep{maes2026lewm}.  \method{} uses physical states to supervise the grounder and common resets to collect intervention branches.  Both methods receive pixels, past actions, and a goal image at deployment.

\tightparagraph{Metrics.}
We report binary task success over 100 paired episodes and the sample mean and standard deviation across three independently trained full-pipeline models.

\subsection{Goal-conditioned control}

\begin{table*}[h]
\centering
\small
\caption{Planning success rate (\%) across four continuous-control environments.  \method{} reports mean $\pm$ sample s.d. over three independent full-pipeline training seeds, each evaluated on the same 100 paired trials.  All comparison rows are reproduced from Table~1 of \citet{zhao2026subjepa}: its LeWM and Sub-JEPA entries average six evaluation seeds with 50 trajectories per seed, whereas its PLDM and DINO-WM entries are transcribed from \citet{maes2026lewm}.  Bold highlights the leading result(s) in each column; on the near-saturated TwoRoom task, both DINO-WM's $100.00$ and \method{}'s $99.0\pm1.0$ are highlighted.}
\label{tab:main}
\resizebox{\textwidth}{!}{%
\begin{tabular}{lcccc}
\toprule
Method & TwoRoom & Reacher & Push-T & OGBench-Cube \\
\midrule
PLDM \citep{sobal2025rewardfree} & 97.00 & 78.00 & 78.00 & 65.00 \\
DINO-WM \citep{zhou2024dinowm} & \textbf{100.00} & 79.00 & 74.00 & 86.00 \\
LeWM \citep{maes2026lewm} & $84.33\pm4.23$ & $82.67\pm4.42$ & $84.67\pm6.53$ & $67.33\pm5.01$ \\
Sub-JEPA \citep{zhao2026subjepa} & $95.00\pm2.76$ & $84.00\pm4.00$ & $\bm{89.00\pm5.33}$ & $76.33\pm5.99$ \\
\textbf{\method{} (ours)} & \textbf{\TwoRoomFIRM} & \textbf{\ReacherFIRM} & \PushTFIRM & \textbf{\CubeFIRM} \\
\bottomrule
\end{tabular}
}
\end{table*}

\begin{figure*}[t]
\centering
\includegraphics[width=0.96\textwidth]{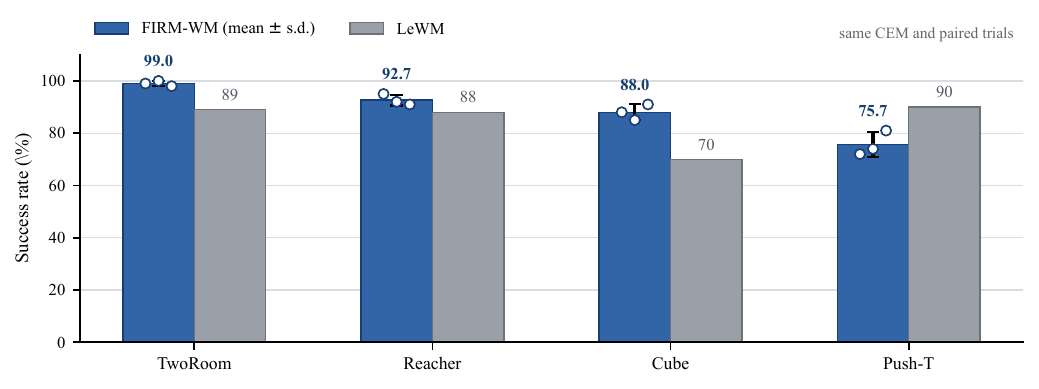}
\caption{\textbf{Control results with optimization variability.} Bars show the mean of three independently trained FIRM-WM seeds; error bars are sample standard deviations and circles are individual seeds.  LeWM uses its fixed checkpoint on the same paired evaluation trials and matched CEM budget.  FIRM improves all three seeds on TwoRoom, Reacher, and Cube; Push-T exposes the persistent-contact boundary.}
\label{fig:control}
\end{figure*}

Table~\ref{tab:main} places \method{} in the broader published landscape, while Figure~\ref{fig:control} reports our matched checkpoint comparison.  The individual \method{} success rates for Seeds~1--3 are 99\%, 100\%, and 98\% on TwoRoom; 95\%, 92\%, and 91\% on Reacher; and 88\%, 85\%, and 91\% on Cube.  The released LeWM checkpoint reaches 89\%, 88\%, and 70\% on the same respective paired trials.  Thus every \method{} seed exceeds the matched checkpoint in these three environments; its mean success rate is 99.0\% on TwoRoom, 92.7\% on Reacher, and 88.0\% on Cube.  The corresponding Push-T seed results are 72\%, 74\%, and 81\%.

Push-T yields \PushTFIRM\% versus \PushTLeWM\% for LeWM, with the same direction across all three independently trained models.  Its sustained object contact couples pose, velocity, and solver mode over the planning horizon, one regime not closed by the present compact deterministic transition.

\subsection{Compactness and planning efficiency}

\begin{table}[!htbp]
\centering
\small
\caption{Active deployment size and recorded planner-loop time on the same evaluation host.  FIRM seconds are mean $\pm$ sample s.d. over three full-pipeline seeds; the LeWM checkpoint is fixed.  Speedup averages the three per-seed ratios.}
\label{tab:efficiency}
\resizebox{\textwidth}{!}{%
\begin{tabular}{lrrrrl}
\toprule
Environment & FIRM active & FIRM sec./ep. & LeWM sec./ep. & Speedup & Matched CEM budget \\
\midrule
TwoRoom & 3.41M & $0.416\pm0.056$ & 3.224 & 7.83$\times$ & 300 / 30 it. / top-30 / $H{=}5$ \\
Reacher & 3.16M & $0.504\pm0.149$ & 5.550 & 11.60$\times$ & 300 / 30 it. / top-30 / $H{=}5$ \\
Push-T & 2.98M & $1.860\pm0.243$ & 3.926 & 2.13$\times$ & 300 / 30 it. / top-30 / $H{=}5$ \\
OGBench-Cube & 3.42M & $0.635\pm0.118$ & 2.389 & 3.84$\times$ & 300 / 10 it. / top-30 / $H{=}5$ \\
\midrule
LeWM reported model size & \multicolumn{4}{l}{$\sim$15M active parameters; time normalized above within each task} & identical per row \\
\bottomrule
\end{tabular}
}
\end{table}

FIRM's active model is 4.4--5.0$\times$ smaller than the approximately 15M-parameter count reported for LeWM.  Among the multi-environment reward-free visual world models in Table~\ref{tab:landscape} with disclosed comparable active counts, FIRM has the smallest active deployment model.

The smaller recurrent state also lowers the cost of evaluating CEM candidates.  Under the matched paired evaluator, the recorded planner loop is 2.13--11.60$\times$ faster across TwoRoom, Reacher, Push-T, and Cube, measured over current/goal encoding, recurrent rollout, CEM population updates, and action selection.

\subsection{Mechanism ablation}

Across Reacher and Cube, the combined factual--interventional pipeline outperforms the best single-source variant in every independently initialized full-pipeline run (Table~\ref{tab:ablation}): \AblFull\% versus \AblFactual\% on Reacher, and \CubeAblFull\% versus \CubeAblIntervention\% on Cube.  Two Reacher-only single-seed diagnostics separately isolate the effects of removing the history fiber (\AblNoHistory\%) and mismatching intervention actions and outcomes (\AblMismatch\%).  Together, the full-chain ablations support complementary factual and common-reset supervision, while the diagnostics support history-dependent state and correct action--outcome correspondence.  Appendix~\ref{app:ablation} and Figure~\ref{fig:ablation} give the protocols and per-run results.

\begin{table}[t]
\centering
\small
\caption{Full-chain training-source ablations.  Values are success (\%), mean $\pm$ sample s.d. over three independently initialized training runs.  $\dagger$ marks Reacher-only single-seed diagnostics; LeWM is the fixed paired baseline.  All entries use released-original CEM.}
\label{tab:ablation}
\begin{tabular}{lcc}
\toprule
Training condition & Reacher & Cube \\
\midrule
Factual only & \AblFactual & \CubeAblFactual \\
Intervention only & \AblIntervention & \CubeAblIntervention \\
Factual + intervention (FIRM) & \textbf{\AblFull} & \textbf{\CubeAblFull} \\
No history fiber$^\dagger$ & \AblNoHistory & -- \\
Mismatched outcomes$^\dagger$ & \AblMismatch & -- \\
\midrule
LeWM (fixed paired baseline) & \AblLeWM & \CubeAblLeWM \\
\bottomrule
\end{tabular}
\end{table}

\section{Limitations and Discussion}

\tightparagraph{Training-time supervision and intervention access.}
FIRM uses physical configuration labels to train $G_\psi$ and environment state-setting interfaces to collect alternative-action trajectories from common resets.  The former is supervision for a learned module; the latter is used only during data collection.  Neither is available to the deployed controller, which uses images, action history, and a goal image.

\tightparagraph{Intervention support and configuration.}
Common-reset interventions calibrate the sampled anchors and actions, while the typed configuration assumes that goal variables and their topology are known.  Broader intervention acquisition and learned goal quotients are natural extensions when either support is unavailable.

\tightparagraph{Persistent contact.}
On Push-T, FIRM averages \PushTFIRM\% versus \PushTLeWM\% for LeWM.  Sustained contact couples geometry, velocity, and solver mode over multiple CEM steps; the result is consistent with a compact deterministic transition being insufficient for this multimodal regime.  Richer conditional dynamics are the direct next extension.

\section{Conclusion}

Reward-free pixel planning requires a state that is both goal-comparable and dynamically sufficient, together with a transition calibrated for the alternative actions queried by CEM.  \method{} supplies these properties through a recurrent factorization into a goal-comparable configuration and a history-dependent dynamic fiber, trained on factual trajectories and common-reset interventions.  Across three independently initialized models, it improves matched LeWM control on TwoRoom, Reacher, and Cube while using fewer active parameters and less recorded planner-loop time.  Push-T reverses the control advantage and locates the remaining challenge in persistent-contact dynamics.  Taken together, the results support state factorization and common-reset interventional calibration as practical design principles for compact reward-free pixel planning.

\bibliography{references}
\bibliographystyle{abbrvnat}

\clearpage
\appendix

\section{Notation Reference}
\label{app:notation}
\FloatBarrier

\begin{table*}[h]
\centering
\scriptsize
\setlength{\tabcolsep}{4pt}
\renewcommand{\arraystretch}{0.90}
\caption{Notation used throughout the paper.  Raw environment time $t$ and local macro index $k$ are distinct.  The symbol $\chi$ denotes training-only auxiliary dynamics, while $d_{\mathcal C}$ denotes the deployment goal cost.}
\label{tab:notation}
\begin{tabular}{@{}p{2.0cm}p{2.35cm}p{9.0cm}@{}}
\toprule
Group & Symbol & Meaning \\
\midrule
\multirow{5}{*}{Environment}
  & $t$ & Raw environment time index. \\
  & $o_t$ & Pixel observation at raw time $t$. \\
  & $a_t\in\mathcal A$ & Raw control applied at time $t$. \\
  & $h_t$ & Causal observation--action history available at time $t$. \\
  & $o^g$ & Goal image. \\
\midrule
\multirow{5}{*}{Macro rollout}
  & $S$ & Number of raw controls in one macro action. \\
  & $k$ & Local macro-step index within a predicted rollout. \\
  & $H$ & Planning horizon in macro steps. \\
  & $u_k$ & Macro action applied at rollout step $k$. \\
  & $U$ & Candidate sequence of $H$ macro actions. \\
\midrule
\multirow{7}{*}{Encoded state}
  & $p_t$ & Spatial visual tokens extracted from $o_t$. \\
  & $b_t$ & Causal history belief formed from recent tokens and actions. \\
  & $s_t=(c_t,m_t)$ & State used by the recurrent transition. \\
  & $c_t$ & Typed, goal-comparable configuration. \\
  & $\delta_t^c$ & Configuration innovation used to initialize the fiber. \\
  & $m_t$ & History-dependent dynamic fiber. \\
  & $\chi_t$ & Environment-typed dynamic target decoded from $m_t$ during training only. \\
\midrule
\multirow{6}{*}{Learned modules}
  & $E_{\phi_E}$ & Pixel encoder that produces spatial tokens. \\
  & $B_{\phi_B}$ & Causal history-belief module. \\
  & $G_\psi$ & Configuration grounder. \\
  & $R_\omega$ & History-to-fiber initializer. \\
  & $T_\theta$ & Recurrent action-conditioned transition. \\
  & $D_\rho$ & Training-only auxiliary dynamics decoder. \\
\midrule
\multirow{10}{*}{Supervision}
  & $x_i^{\mathrm{sim}}$ & Recorded environment state at intervention anchor $i$. \\
  & $\DF$ & Factual rollout dataset. \\
  & $\DI$ & Common-reset intervention rollout dataset. \\
  & $\xi^{\mathrm I}_{ij}$ & Intervention branch $j$ from anchor $i$. \\
  & $K_i$ & Number of intervention branches at anchor $i$. \\
  & $\sigma\in\{\mathrm F,\mathrm I\}$ & Data-source index: factual or interventional. \\
  & $\mathcal A_{e,r}$ & Active data sources in recurrent stage $r$ of environment $e$. \\
  & $R_e$ & Number of recurrent stages for environment $e$. \\
  & $N_{e,r}$ & Number of updates in recurrent stage $r$ of environment $e$. \\
  & $N_{\mathrm{rec}}^{(e)}$ & Total recurrent-transition updates for environment $e$. \\
\midrule
\multirow{6}{*}{Loss}
  & $\mathcal B$ & Training minibatch. \\
  & $\Lambda_\kappa$ & Topology-aware configuration displacement map. \\
  & $q(k)$ & Predecessor index used by the recursive loss. \\
  & $\bar c_k$ & Anchored configuration target at macro step $k$. \\
  & $\mathcal E(\mathcal B)$ & Multiset of per-example tail errors in minibatch $\mathcal B$. \\
  & $\ell_\sigma$ & Training loss for source $\sigma$. \\
\midrule
\multirow{2}{*}{Planning}
  & $d_{\mathcal C}$ & Squared terminal distance in configuration space. \\
  & $J$ & Objective assigned to a candidate action sequence. \\
\bottomrule
\end{tabular}
\end{table*}
\FloatBarrier

\section{Task Interfaces and Model Geometry}
\label{app:interfaces}
\FloatBarrier

The recurrent transition evolves in macro time.  If one macro action contains $S$ raw controls, then
\begin{equation}
    u_k=(a_{t+Sk},\ldots,a_{t+S(k+1)-1})\in\mathcal A^S,
    \qquad U=(u_0,\ldots,u_{H-1}).
    \label{eq:macroaction}
\end{equation}
All experiments use $S=5$ and $H=5$, so each candidate sequence spans 25 raw controls.  Table~\ref{tab:plannercontract} gives the remaining CEM contract.

For configurations with $N_\circ$ periodic scalar variables, write
\begin{equation}
c=\left(e(\vartheta_1),\ldots,e(\vartheta_{N_\circ}),c^{\mathrm{euc}}\right),
\qquad e(\vartheta)=(\sin\vartheta,\cos\vartheta),
\end{equation}
where $c^{\mathrm{euc}}$ collects the non-periodic Euclidean coordinates of the task configuration.  We then define
\begin{equation}
d_{\mathcal C}(c,c')=
\|c^{\mathrm{euc}}-c^{\prime\mathrm{euc}}\|_2^2
+\sum_{\ell=1}^{N_\circ}\|e(\vartheta_\ell)-e(\vartheta'_\ell)\|_2^2.
\label{eq:configuration-cost}
\end{equation}
The periodic contribution in Equation~\eqref{eq:configuration-cost} is the squared chord distance between two unit-circle encodings.  Let $\delta_\ell=\operatorname{wrap}(\vartheta_\ell-\vartheta'_\ell)\in[-\pi,\pi)$.  Then $\|e(\vartheta_\ell)-e(\vartheta'_\ell)\|_2^2=4\sin^2(\delta_\ell/2)$, so equivalent angles on opposite sides of the $\pm\pi$ boundary remain close.  TwoRoom and Cube have no periodic coordinates; the term is applied to both Reacher joints and to the Push-T block orientation.

\begin{table*}[h]
\centering
\small
\caption{Typed interfaces used by the shared configuration--fiber transition.  History length is measured in observations; every macro action contains five raw environment controls.}
\label{tab:interfaces}
\resizebox{\textwidth}{!}{%
\begin{tabular}{lclcccc}
\toprule
Environment & $\dim(c)$ & Goal-comparable configuration & Periodic pairs & $\dim(\chi)$ & History frames & $\dim(u_k)$ \\
\midrule
TwoRoom & 2 & agent position & 0 & 2 & 3 & 10 \\
Reacher & 4 & sine/cosine of two joint angles & 2 & 2 & 3 & 10 \\
Push-T & 6 & block orientation and agent/block positions & 1 & 5 & 3 & 10 \\
OGBench-Cube & 3 & fixed affine-normalized block position in 3-D & 0 & 25 & 3 & 25 \\
\bottomrule
\end{tabular}%
}
\end{table*}

\begin{table*}[h]
\centering
\small
\caption{Environment-specific configuration and terminal-cost instantiations.  Here $p^{\mathrm a}$ denotes agent position, $p^{\mathrm b}$ denotes block position, and a subscript $g$ denotes the value inferred from the goal image.  One periodic scalar is stored as one unit-circle pair $e(\vartheta)=(\sin\vartheta,\cos\vartheta)$; it does not introduce a second physical degree of freedom.}
\label{tab:configuration-costs}
\begin{tabular}{@{}lp{3.8cm}cp{5.9cm}@{}}
\toprule
Environment & Configuration $c$ & $N_\circ$ & Terminal cost \\
\midrule
TwoRoom
& $c=p^{\mathrm a}=(x^{\mathrm a},y^{\mathrm a})$
& 0
& $\|p^{\mathrm a}-p_g^{\mathrm a}\|_2^2$ \\
Reacher
& $c=(e(\vartheta_1),e(\vartheta_2))$
& 2
& $\sum_{j=1}^{2}\|e(\vartheta_j)-e(\vartheta_{g,j})\|_2^2$ \\
Push-T
& $c=(e(\vartheta^{\mathrm b}),p^{\mathrm a},p^{\mathrm b})$
& 1
& $\max\{r_\vartheta,r_p\}^{2}$, where
  $r_\vartheta=|\operatorname{wrap}(\vartheta^{\mathrm b}-\vartheta_g^{\mathrm b})|/\tau_\vartheta$
  and $r_p=\|\Delta p\|_2/\tau_p$ (task-normalized) \\
OGBench-Cube
& fixed affine-normalized 3-D block position
& 0
& $\|c-c_g\|_2^2$ \\
\bottomrule
\end{tabular}
\end{table*}

Push-T uses the task's fixed orientation and position tolerances to put its two error types on a common scale.  Define the concatenated position error by $\Delta p=[(p^{\mathrm a}-p_g^{\mathrm a})^\top,(p^{\mathrm b}-p_g^{\mathrm b})^\top]^\top$.  The normalized errors are $r_\vartheta=|\operatorname{wrap}(\vartheta^{\mathrm b}-\vartheta_g^{\mathrm b})|/(\pi/9)$ for block orientation and $r_p=\|\Delta p\|_2/(20/256)$ for agent--block position in normalized image coordinates.  Its terminal cost $\max\{r_\vartheta,r_p\}^2$ is small only when both errors approach their respective tolerances.  The other three tasks use the corresponding instance of Equation~\eqref{eq:configuration-cost}; for TwoRoom and Cube, the periodic sum is empty.  The grounder and recurrent transition normalize every predicted sine--cosine block to unit length before the cost is evaluated.

The dynamic target is used only to shape the recurrent fiber.  It contains configuration velocity or the remaining task state after removing the static goal configuration.  In Cube, for example, the target includes robot joint position and velocity, end-effector pose, gripper state, contact indicator, and block orientation, while the terminal cost uses only affine-normalized block position.  These simulator quantities supervise $D_\rho$ during training but are not read by the controller.  At deployment, the current image produces $c_t=G_\psi(E_{\phi_E}(o_t,\mathbf 0))$, the goal image produces $c_g=G_\psi(E_{\phi_E}(o^g,\mathbf 0))$, and CEM compares the recursively predicted $\hat c_H$ with $c_g$.  Configuration is therefore inferred from pixels rather than measured from the simulator.

Concretely, TwoRoom uses normalized agent position for $c_t$ and its finite displacement for $\chi_t$.  Reacher maps both joint angles to sine--cosine pairs and uses the two joint velocities as $\chi_t$.  The Push-T interface concatenates block-angle sine--cosine with normalized agent and block positions; its five-dimensional auxiliary target is the macro-step change in agent position, block position, and wrapped block angle.  Cube uses a fixed affine-normalized three-dimensional block position for $c_t$ and a 25-dimensional auxiliary target $\chi_t$: six arm joint positions, six joint velocities, normalized 3-D end-effector position, end-effector yaw as a sine--cosine pair, gripper opening, gripper contact, and a sign-invariant 6-D block-rotation chart.  These configuration coordinates use task-fixed transforms rather than data-estimated standardizers.  Separately, model-side action statistics and, where used, auxiliary-dynamics statistics are estimated from the training split and frozen in the checkpoint; deployment reads those stored constants and does not query simulator state.

The spatial tokenizer used in TwoRoom and Cube has four convolutional stages with channels $32,64,96,64$, GroupNorm and GELU activations, and a learned $7\times7$ positional embedding.  Push-T uses the same channel backbone with a $14\times14$ typed spatial output and a grid-preserving grounder.  Training-only decoders and change-support heads are excluded from active deployment counts.  Reacher uses a compact geometry-oriented pixel encoder and a causal history filter.  Across all tasks, FIRM-WM uses three observations, a 128-dimensional recurrent fiber, a 384-dimensional transition hidden state, one recurrent transition, five macro rollout steps, and a typed configuration update.  The Push-T transition places an action-conditioned spatial residual inside the shared hidden update; it does not change the recurrent state, output heads, or planner interface.

\begin{table*}[h]
\centering
\small
\caption{Released-original-CEM evaluation contract.  Candidate action coordinates use the train-split standardizer stored in each checkpoint, search Gaussians are unbounded, and the deployed sequence is the final elite mean.  No row changes the solver between FIRM-WM and LeWM.}
\label{tab:plannercontract}
\begin{tabular}{lrrrrrr}
\toprule
Environment & Candidates & Iterations & Elites & Macro horizon & Raw steps / macro & Goal offset \\
\midrule
TwoRoom & 300 & 30 & 30 & 5 & 5 & 25 \\
Reacher & 300 & 30 & 30 & 5 & 5 & 25 \\
Push-T & 300 & 30 & 30 & 5 & 5 & 25 \\
OGBench-Cube & 300 & 10 & 30 & 5 & 5 & 25 \\
\bottomrule
\end{tabular}
\end{table*}

\FloatBarrier
\section{Module-wise Parameter Accounting}

\begin{table*}[h]
\centering
\small
\caption{Active deployment parameters by module.  Pixel + history frontend combines the deployed visual encoder/tokenizer and causal history module.  Training-only reconstruction and change-support heads are omitted.}
\label{tab:params}
\resizebox{\textwidth}{!}{%
\begin{tabular}{lrrrr}
\toprule
Environment & Pixel + history frontend & Configuration grounder & Recurrent transition & Active total \\
\midrule
TwoRoom & 0.374M & 1.613M & 1.424M & 3.411M \\
Reacher & 0.992M & 0.828M & 1.343M & 3.162M \\
Push-T & 0.449M & 0.157M & 2.371M & 2.977M \\
Cube & 0.376M & 1.613M & 1.434M & 3.424M \\
\bottomrule
\end{tabular}
}
\vspace{2pt}
\parbox{0.96\textwidth}{\footnotesize Totals are direct sums of active modules and are rounded only for display.}
\end{table*}
\FloatBarrier

\section{Training and Evaluation Disclosure}
\label{app:disclosure}

\paragraph{Data sets.}
The offline training and validation splits contain 9,000/1,000 episodes for TwoRoom, Reacher, and Cube, and 16,685/2,000 episodes for Push-T.  Table~\ref{tab:provenance} gives the factual rollout segments and intervention data used by the final joint recurrent stage.  The same raw intervention records may be shared across independent training seeds, but each seed encodes them with its independently trained frontend and learns a separate transition.

\paragraph{Anchor and branch collection.}
An anchor is one eligible time index sampled from a training episode; anchors use distinct source episodes.  The index must leave enough preceding observations to form the three-frame history $h_i$ and enough remaining steps for a 25-control rollout.  For TwoRoom, Push-T, and Cube, one branch replays the recorded 25-action suffix and the remaining branches use Gaussian action proposals clipped to the action bounds.  The proposal statistics come from the training split for TwoRoom and Cube and from the complete offline action table for Push-T.  Reacher samples all 16 branches uniformly from its action bounds.  Every branch runs for 25 raw controls and records targets after controls 5, 10, 15, 20, and 25, giving $H=5$ macro steps.  Anchor selection and branch sampling do not use reward, success, goal distance, or planner rank.

\paragraph{Common-reset intervention collection.}
The collector does not serialize the simulator process.  Before each intervention branch, we reset the environment and restore the same recorded values $x_i^{\mathrm{sim}}$ of the variables accepted by the task's state-setting interface.  For TwoRoom this condition is the two-dimensional agent position.  Reacher restores the two joint positions and velocities, sets the target geometry, and updates the derived physics state.  Push-T installs its seven-dimensional public state: agent position, block position and orientation, and agent velocity; other simulator variables take the environment's reset values.  Cube passes the 21-dimensional MuJoCo position and 20-dimensional velocity vectors through the released reset interface and sets the target block pose; the solver warm-start state is reinitialized.  Thus ``common reset'' means that the state variables exposed by the released environment interface are set to identical values for all branches at an anchor, not that the complete simulator memory image is saved and restored.

\raggedbottom
\begin{table}[H]
\centering
\scriptsize
\setlength{\tabcolsep}{2.5pt}
\renewcommand{\arraystretch}{0.95}
\caption{Common-reset factual-suffix replay on training anchors.  Errors are configuration discrepancies in task success-tolerance units after 25 controls ($H{=}5$); ``Max'' is over all five macro horizons.  The fixed gate requires P95 $\leq0.25$ at every horizon and Max $\leq1.0$.  This tests restored-interface fidelity, not identity of hidden simulator or solver state.}
\label{tab:reset-replay}
\begin{tabular}{@{}lrrrrc@{}}
\toprule
Environment & Anchors & H5 mean & H5 P95 & Max & Gate \\
\midrule
TwoRoom & 128 & 0 & 0 & 0 & Pass \\
Reacher & 442 & $4.52{\times}10^{-15}$ & $1.75{\times}10^{-14}$ & $3.55{\times}10^{-14}$ & Pass \\
Push-T & 512 & 0.0613 & 0.1424 & 0.9548 & Pass \\
OGBench-Cube & 128 & 0.00352 & 0.01035 & 0.06111 & Pass \\
\bottomrule
\end{tabular}
\vspace{1pt}
\parbox{0.98\columnwidth}{\footnotesize Reacher includes the 442 of 512 prespecified anchors with a complete 25-control factual suffix; the other 70 are excluded solely by episode length and are not replaced.}
\end{table}

\paragraph{Training protocol.}
Each seed independently initializes and trains the task-specific frontend, configuration grounder, and recurrent transition under the same fixed data split, intervention collections, update schedule, optimizer settings, and final-checkpoint rule.  Algorithm~\ref{alg:training} covers all updates to the recurrent modules.  Table~\ref{tab:recurrent-schedule} gives the complete sequence.  The frontend and grounder are fixed within each recurrent stage but may be refined between stages; recurrent parameters are carried forward.  If a stage enlarges the recurrent input, as in Push-T, existing recurrent weights are retained and weights on the new input coordinates are initialized to zero.

\begin{table}[!t]
\centering
\scriptsize
\setlength{\tabcolsep}{1.5pt}
\renewcommand{\arraystretch}{0.90}
\caption{Module training contracts.  Arrows denote consecutive stages; update counts and learning rates align from left to right.  For OGBench-Cube, S1--S3 denote initial module training, joint Tokenizer--Grounder refinement, and height-balanced refinement.}
\label{tab:module-training}
\begin{tabular}{@{}p{1.20cm}p{1.80cm}p{3.50cm}p{2.10cm}p{1.55cm}p{1.30cm}@{}}
\toprule
Module & Initialization & Objective & Updates & LR ($\times10^{-4}$) & Status \\
\midrule
\multicolumn{6}{@{}l}{\textbf{TwoRoom}} \\
Tokenizer & random & RGB reconstruction + change support & 100 & $2$ & frozen \\
Grounder & random head & configuration + finite-difference regression & 1,000 & $3$ & frozen \\
History module & random & causal token-dynamics NLL & 500 & $3$ & frozen \\
Unified transition & random $\rightarrow$ carried & recursive $\mathrm F\rightarrow(\mathrm F+\mathrm I)$ rollout & $3{,}000\rightarrow3{,}000$ & $3\rightarrow1$ & active \\
\midrule
\multicolumn{6}{@{}l}{\textbf{Reacher}} \\
Pixel encoder & random & reconstruction + equivariance + geometry coverage & $500\rightarrow2{,}000$ & $2\rightarrow0.5$ & frozen \\
Grounder & staged heads & physical, configuration, horizon, residual regression & $1{,}500\rightarrow2{,}000\rightarrow3{,}000\rightarrow3{,}000$ & \begin{tabular}[t]{@{}l@{}}$2;\ 0.5/5;$\\$2;\ 1/5$\end{tabular} & frozen \\
History module & random & causal representation dynamics & 500 & $2$ & frozen \\
Unified transition & random $\rightarrow$ carried & recursive $\mathrm I\rightarrow\mathrm I\rightarrow(\mathrm F+\mathrm I)$ rollout & $4{,}000\rightarrow2{,}000\rightarrow3{,}000$ & $3\rightarrow1\rightarrow1$ & active \\
\midrule
\multicolumn{6}{@{}l}{\textbf{Push-T}} \\
Tokenizer & pretrained frontend & position + visual preservation & 500 & $1$ & frozen \\
Grounder & random head & typed configuration regression & 2,000 & $3$ & frozen \\
History module & inherited parent & rollout objective & 2,000 & $0.05$ & joint; active \\
Unified transition & recurrent parent & factual + common-reset + structured rollout & 2,000 & $0.3$ & joint; active \\
\midrule
\multicolumn{6}{@{}l}{\textbf{OGBench-Cube}} \\
Tokenizer & random & reconstruction + change support + token anchoring & \begin{tabular}[t]{@{}l@{}}S1: 500\\S2: 1,000\\S3: 2,000\end{tabular} & \begin{tabular}[t]{@{}l@{}}S1: 2\\S2: 0.1\\S3: 0.05\end{tabular} & frozen \\
Grounder & random head & pose regression + height-balanced tail & \begin{tabular}[t]{@{}l@{}}S1: 1,500\\S2: 1,000\\S3: 2,000\end{tabular} & \begin{tabular}[t]{@{}l@{}}S1: 3\\S2: 1\\S3: 0.5\end{tabular} & frozen \\
History module & random & causal token-dynamics NLL & 1,000 & $3$ & frozen \\
Unified transition & random & recursive $(\mathrm F+\mathrm I)$ rollout & 3,000 & $1$ & active \\
\bottomrule
\end{tabular}
\vspace{2pt}

\parbox{0.90\linewidth}{\scriptsize All stages use AdamW.  Frontend and grounder checkpoints use held-out representation or configuration gates; recurrent checkpoints use held-out multi-horizon rollout gates.}
\end{table}

\begin{table*}[h]
\centering
\small
\caption{Complete recurrent-transition update schedules.  Arrows separate consecutive stages; $\mathrm F$ and $\mathrm I$ denote factual and intervention sources.  The final stage in every row uses Equation~\eqref{eq:total}.}
\label{tab:recurrent-schedule}
\resizebox{0.92\textwidth}{!}{%
\begin{tabular}{lccc}
\toprule
Environment & Active sources by stage & Updates by stage & $N_{\mathrm{rec}}^{(e)}$ \\
\midrule
TwoRoom & $\mathrm F\rightarrow(\mathrm F+\mathrm I)$ & $3{,}000\rightarrow3{,}000$ & $6{,}000$ \\
Reacher & $\mathrm I\rightarrow\mathrm I\rightarrow(\mathrm F+\mathrm I)$ & $4{,}000\rightarrow2{,}000\rightarrow3{,}000$ & $9{,}000$ \\
Push-T & $(\mathrm F+\mathrm I)\times6\rightarrow\mathrm{SR}$ & $5{,}000\rightarrow3{,}000\rightarrow3{,}000\rightarrow5{,}000\rightarrow10{,}000\rightarrow3{,}000\rightarrow2{,}000$ & $31{,}000$ \\
OGBench-Cube & $\mathrm F+\mathrm I$ & $3{,}000$ & $3{,}000$ \\
\bottomrule
\end{tabular}%
}
\end{table*}

For Push-T, the task-typed Tokenizer and Grounder are refined for 500 and 2,000 updates, respectively, before the final 2,000-update unified-transition refinement.

\paragraph{Planning.}
All main results use 300 CEM samples, top 30 elites, five macro steps, receding horizon five, and five raw environment steps per macro action.  Search occurs in action coordinates standardized by the training-split mean and scale stored in the checkpoint, is unbounded, and returns the final elite mean.  TwoRoom, Reacher, and Push-T use 30 CEM iterations; Cube follows the 10-iteration LeWM protocol \citep{maes2026lewm}.

\section{Complete Supervision and Training Procedure}
\label{app:supervision}

For either a factual rollout segment or an intervention branch, the target at macro index $k$ is read after exactly $Sk=5k$ raw simulator controls.  Define the topology-aware relative-configuration map
\begin{equation}
\Gamma_{\mathcal C}(c',c)=
\left(c^{\prime\mathrm{euc}}-c^{\mathrm{euc}},
e(\vartheta'_1-\vartheta_1),\ldots,
e(\vartheta'_{N_\circ}-\vartheta_{N_\circ})\right),
\label{eq:relative-configuration}
\end{equation}
with $e(\vartheta)=(\sin\vartheta,\cos\vartheta)$.  To represent the change from a reference configuration $c$ to a later configuration $c'$, define the displacement operator
\begin{equation}
\Lambda_\kappa(c',c)=
\begin{cases}
c'-c, & \kappa=\mathrm{emb},\\
\Gamma_{\mathcal C}(c',c), & \kappa=\mathrm{rel}.
\end{cases}
\label{eq:displacement-operator}
\end{equation}
The map $\Lambda_\kappa$ is deterministic rather than learned.  For $\kappa=\mathrm{emb}$, it subtracts the stored configuration coordinates directly.  For $\kappa=\mathrm{rel}$, $\Gamma_{\mathcal C}$ subtracts the Euclidean coordinates and composes each periodic angle before applying its sine--cosine encoding, avoiding a discontinuity at angle wrap-around.  The displacement loss applies the same map to predicted and target configuration pairs relative to the predecessor $q(k)$.  For a finite multiset $\mathcal E$ of scalar errors, define
\begin{equation}
\operatorname{TopMean}_{0.25}(\mathcal E)
=\frac{1}{\lceil0.25|\mathcal E|\rceil}
\sum_{z\in\operatorname{Top}_{\lceil0.25|\mathcal E|\rceil}(\mathcal E)}z,
\label{eq:topmean}
\end{equation}
where $\operatorname{Top}_r(\mathcal E)$ returns the $r$ largest elements with multiplicity.  For each minibatch $\mathcal B$, $\mathcal L_{\mathrm{tail}}(\mathcal B)=\operatorname{TopMean}_{0.25}(\mathcal E(\mathcal B))$.  The error multiset and component weights are specified separately for each source and task.  Table~\ref{tab:loss-contract} gives the complete contract.  ``Terminal configuration MSE'' means one scalar per rollout record at $k=H$; the Reacher tail instead contributes one maximum joint-chord error per record and supervised macro index.  The same decoder $D_\rho$ is applied at $k=0$ and at future steps; there is no separate initial-state decoder.

For a minibatch $\mathcal B=\{\xi_n\}_{n=1}^{N}$, predictions are generated by Equation~\eqref{eq:closedtransition}; batch indices are suppressed below.  Write $\operatorname{MSE}(X,Y)$ for the mean squared error over every scalar entry of equally shaped tensors.  Here $\hat c_0=G_\psi(E_{\phi_E}(o_t,\mathbf 0))$ is the configuration inferred from the current image, whereas $c_0$ is the ground-truth task configuration at the same time step, obtained from the recorded simulator state through the task-specific coordinate map in Appendix~\ref{app:interfaces}.  Define the anchored target by $\bar c_0=\hat c_0$ and $\bar c_k=c_k$ for $k\geq1$.  The first displacement comparison therefore uses the same inferred anchor $\hat c_0$ on its predicted and target sides, preventing current-frame grounding error from being counted again as transition-increment error.  The exact components grouped conceptually in Equation~\eqref{eq:loss-groups} are
\begin{align}
\mathcal L_c(\mathcal B) &= \operatorname{MSE}(\hat c_{1:H},c_{1:H}), \nonumber\\
\mathcal L_{\mathrm{disp}}(\mathcal B) &= \operatorname{MSE}\!\left(
 \{\Lambda_\kappa(\hat c_k,\hat c_{q(k)})\}_{k=1}^{H},
 \{\Lambda_\kappa(c_k,\bar c_{q(k)})\}_{k=1}^{H}\right), \nonumber\\
\mathcal L_\chi(\mathcal B) &= \operatorname{MSE}\!\left(
 \{D_\rho(\hat m_k)\}_{k=1}^{H},\chi_{1:H}\right),
 \qquad
\mathcal L_{\chi,0}(\mathcal B) = \operatorname{MSE}\!\left(D_\rho(\hat m_0),\chi_0\right).
\label{eq:loss-components}
\end{align}
For source $\sigma\in\{\mathrm F,\mathrm I\}$, the training loss is
\begin{equation}
\ell_\sigma(\mathcal B)=
w_{\sigma,c}\mathcal L_c+w_{\sigma,\mathrm{disp}}\mathcal L_{\mathrm{disp}}
+w_{\sigma,\mathrm{tail}}\mathcal L_{\mathrm{tail}}
+w_{\sigma,\chi}\mathcal L_\chi+w_{\sigma,\chi0}\mathcal L_{\chi,0}.
\label{eq:loss}
\end{equation}
In the three-part presentation of Equation~\eqref{eq:loss-groups}, the future-configuration group contains $\mathcal L_c$ and $\mathcal L_{\mathrm{tail}}$, the configuration-displacement group contains $\mathcal L_{\mathrm{disp}}$, and the auxiliary-dynamics group contains $\mathcal L_\chi$ and $\mathcal L_{\chi,0}$.

\begin{table*}[h]
\centering
\scriptsize
\caption{Exact rollout-loss contract.  Each row lists $(w_{\sigma,c},w_{\sigma,\mathrm{disp}},w_{\sigma,\mathrm{tail}},w_{\sigma,\chi},w_{\sigma,\chi0})$ from Equation~\eqref{eq:loss}.}
\label{tab:loss-contract}
\resizebox{\textwidth}{!}{%
\begin{tabular}{lllll}
\toprule
Task & Source $\sigma$ & $(\kappa,q(k))$ & Error multiset $\mathcal E(\mathcal B)$ for $\mathcal L_{\mathrm{tail}}$ & component weights \\
\midrule
TwoRoom & factual $\mathrm F$ & $(\mathrm{emb},k-1)$ & terminal configuration MSE & $(2,2,0.5,0.5,0.5)$ \\
TwoRoom & intervention $\mathrm I$ & $(\mathrm{emb},k-1)$ & terminal configuration MSE & $(2,2,1.0,0.5,0)$ \\
Reacher & factual $\mathrm F$ & $(\mathrm{rel},0)$ & all-horizon maximum joint chord error & $(2,2,0.5,0.5,0.5)$ \\
Reacher & intervention $\mathrm I$ & $(\mathrm{rel},0)$ & all-horizon maximum joint chord error & $(2,2,0.5,0.5,0.5)$ \\
Push-T & factual $\mathrm F$ & $(\mathrm{emb},k-1)$ & terminal configuration MSE & $(2,2,0.5,0.5,0.5)$ \\
Push-T & intervention $\mathrm I$ & $(\mathrm{emb},k-1)$ & terminal configuration MSE & $(2,2,1.0,0.5,0)$ \\
Push-T & structured refinement $\mathrm{SR}$ & $(\mathrm{emb},k-1)$ & terminal configuration MSE & $(2,2,1.0,0.5,0.5)$ \\
OGBench-Cube & factual $\mathrm F$ & $(\mathrm{emb},k-1)$ & terminal configuration MSE & $(2,2,0.5,0.5,0.5)$ \\
OGBench-Cube & intervention $\mathrm I$ & $(\mathrm{emb},k-1)$ & terminal configuration MSE & $(2,2,1.0,0.5,0)$ \\
\bottomrule
\end{tabular}%
}
\end{table*}

The Push-T structured-refinement minibatch mixes broad factual, IID common-reset, and contact-balanced structured records with source weights $1.0$, $0.5$, and $1.0$.  Contact metadata affects sampling only; reward, success, planner cost, and candidate rank are not training targets.

\FloatBarrier

\paragraph{Cube public reproduction.}
Separate from the three-seed evaluation on 100 paired trials, the public seed-42 50-trial Cube list gives \CubePublicFIRM{} success for FIRM and \CubePublicLeWM{} for LeWM with 300/10/top-30 CEM.  This reproduction is not averaged into Table~\ref{tab:main}.

\section{Full-Chain Training-Source Ablations}
\label{app:ablation}

\begin{figure*}[t]
\centering
\includegraphics[width=0.96\textwidth]{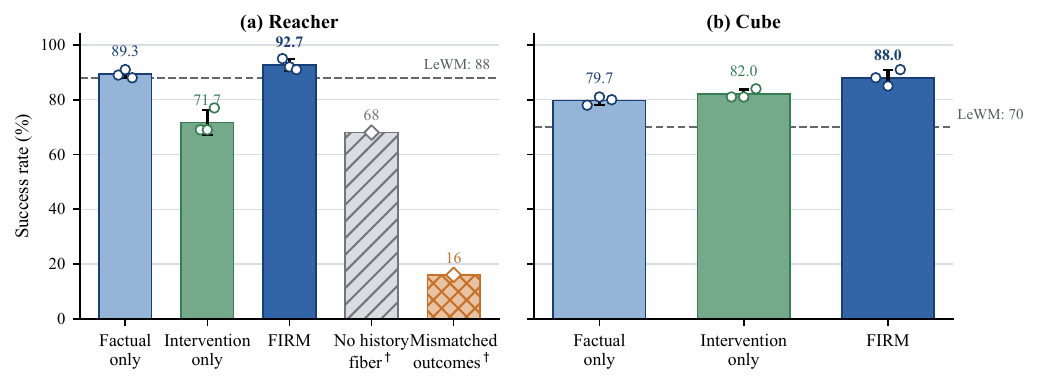}
\caption{\textbf{Full-chain training-source ablations.} Each panel shows three independently initialized training runs (dots), their mean, and sample s.d. for factual-only, intervention-only, and combined training.  The combined pipeline performs best in every run on both tasks.  Reacher additionally includes two separate single-seed diagnostics marked $\dagger$.  All results use released-original CEM; dashed lines show the task-specific fixed LeWM paired baselines.}
\label{fig:ablation}
\end{figure*}

The first three rows of Table~\ref{tab:ablation} retrain the complete Reacher and Cube task pipelines from independent initialization under factual-only, intervention-only, or combined supervision.  Within each environment, each variant uses three random seeds while holding the architecture, stage-wise optimization schedule, data split, and paired evaluation trials fixed.  Figure~\ref{fig:ablation} shows the distributions across independently initialized training runs.  The two additional single-seed diagnostics are Reacher-only; the mismatched-outcome diagnostic preserves the marginal sets of histories, actions, and outcomes while destroying their correspondence.

\section{Full Per-Seed Results}

\begin{table*}[h]
\centering
\scriptsize
\caption{Per-seed paired results.  $\Delta$ is FIRM minus LeWM in success points; intervals are 95\% paired bootstrap intervals.  Rescue/harm count trials won only by FIRM / only by LeWM.  LeWM timing is fixed within an environment because its checkpoint and paired evaluator are reused.}
\label{tab:fullseeds}
\begin{tabular}{llrrrrrr}
\toprule
Environment & Seed & FIRM & LeWM & $\Delta$ [95\% CI] & Rescue / harm & FIRM sec. & LeWM sec. \\
\midrule
\multirow{3}{*}{TwoRoom} & 1 & 99 & 89 & $+10$ [5,16] & 10 / 0 & 0.480 & 3.224 \\
 & 2 & 100 & 89 & $+11$ [5,18] & 11 / 0 & 0.378 & 3.224 \\
 & 3 & 98 & 89 & $+9$ [3,16] & 10 / 1 & 0.391 & 3.224 \\
\midrule
\multirow{3}{*}{Reacher} & 1 & 95 & 88 & $+7$ [0,15] & 11 / 4 & 0.675 & 5.550 \\
 & 2 & 92 & 88 & $+4$ [$-4$,12] & 10 / 6 & 0.407 & 5.550 \\
 & 3 & 91 & 88 & $+3$ [$-5$,11] & 10 / 7 & 0.429 & 5.550 \\
\midrule
\multirow{3}{*}{Cube} & 1 & 88 & 70 & $+18$ [8,28] & 23 / 5 & 0.771 & 2.389 \\
 & 2 & 85 & 70 & $+15$ [5,25] & 21 / 6 & 0.575 & 2.389 \\
 & 3 & 91 & 70 & $+21$ [13,30] & 22 / 1 & 0.560 & 2.389 \\
\midrule
\multirow{3}{*}{Push-T} & 1 & 72 & 90 & $-18$ [$-27$,$-9$] & 3 / 21 & 1.656 & 3.926 \\
 & 2 & 74 & 90 & $-16$ [$-25$,$-7$] & 4 / 20 & 1.795 & 3.926 \\
 & 3 & 81 & 90 & $-9$ [$-18$,0] & 7 / 16 & 2.128 & 3.926 \\
\bottomrule
\end{tabular}
\end{table*}
\FloatBarrier

All TwoRoom and Cube intervals are positive.  Reacher favors FIRM in every seed, with two intervals crossing zero.  Push-T consistently favors LeWM across all three random seeds, separating a persistent-contact limitation from random-seed variation.

All values in Tables~\ref{tab:efficiency} and~\ref{tab:fullseeds} come from the paired evaluator on one local evaluation host equipped with an NVIDIA GeForce RTX 3060 Laptop GPU (6\,GB).  A timer brackets repeated online planner work: current/goal encoding required by the evaluator, recurrent candidate rollout, CEM population updates, and action selection.  It excludes offline cache construction, dataset loading, training, process launch, and simulator creation outside the repeated boundary.  Both methods use the same paired evaluation trials and CEM budget in each task.

\section{Data and Evaluation Protocol}

Table~\ref{tab:provenance} summarizes the data used in each environment.  We train three complete models per environment from independent random initializations while holding the data split, intervention set, and hyperparameters fixed across seeds.

\begin{table*}[h]
\centering
\small
\caption{Dataset statistics for the final joint recurrent stage.  A factual rollout segment contains a three-frame history and its recorded 25-control continuation.  Intervention states are the recorded interface states used to initialize action branches; total branch rollouts equal intervention states times action branches per state.  All three models in an environment are evaluated on the same 100 paired trials.}
\label{tab:provenance}
\resizebox{\textwidth}{!}{%
\begin{tabular}{lrrrrrrl}
\toprule
Environment & Training episodes & Validation episodes & \shortstack{Factual rollout\\segments} & \shortstack{Intervention\\states} & \shortstack{Action branches\\per state} & \shortstack{Branch\\rollouts} & Evaluation trials \\
\midrule
TwoRoom & 9,000 & 1,000 & 16,384 & 128 & 32 & 4,096 & 100 paired trials \\
Reacher & 9,000 & 1,000 & 16,384 & 512 & 16 & 8,192 & 100 paired trials \\
OGBench-Cube & 9,000 & 1,000 & 4,096 & 128 & 16 & 2,048 & 100 paired trials \\
Push-T & 16,685 & 2,000 & 16,384 & 512 & 16 & 8,192 & 100 paired trials \\
\bottomrule
\end{tabular}
}
\end{table*}

\section{LLM Usage Statement}

\small
A large language model was used as a writing assistant to help organize the manuscript, improve prose, and check consistency between claims and project artifacts.  It did not generate experimental measurements.  The authors are responsible for verifying every equation, citation, result, and scientific claim in the submitted version.

\end{document}

%% file: math_commands.tex
\usepackage{amsmath,amsfonts,bm}

\def\eqref#1{equation~\ref{#1}}

\def\1{\bm{1}}

\DeclareMathAlphabet{\mathsfit}{\encodingdefault}{\sfdefault}{m}{sl}
\SetMathAlphabet{\mathsfit}{bold}{\encodingdefault}{\sfdefault}{bx}{n}



%% file: numbers.tex
\newcommand{\TwoRoomFIRM}{99.0\,$\pm$\,1.0}
\newcommand{\TwoRoomLeWM}{89.0}
\newcommand{\ReacherFIRM}{92.7\,$\pm$\,2.1}
\newcommand{\ReacherLeWM}{88.0}
\newcommand{\CubeFIRM}{88.0\,$\pm$\,3.0}
\newcommand{\CubeLeWM}{70.0}

\newcommand{\PushTFIRM}{75.7\,$\pm$\,4.7}
\newcommand{\PushTLeWM}{90.0}

\newcommand{\FIRMParamsMin}{2.98}
\newcommand{\FIRMParamsMax}{3.42}
\newcommand{\LeWMPaperParams}{15}

\newcommand{\AblFactual}{89.3\,$\pm$\,1.5}
\newcommand{\AblIntervention}{71.7\,$\pm$\,4.6}
\newcommand{\AblFull}{92.7\,$\pm$\,2.1}
\newcommand{\AblNoHistory}{68.0}
\newcommand{\AblMismatch}{16.0}
\newcommand{\AblLeWM}{88.0}
\newcommand{\CubeAblFactual}{79.7\,$\pm$\,1.5}
\newcommand{\CubeAblIntervention}{82.0\,$\pm$\,1.7}
\newcommand{\CubeAblFull}{88.0\,$\pm$\,3.0}
\newcommand{\CubeAblLeWM}{70.0}

\newcommand{\CubePublicFIRM}{42/50}
\newcommand{\CubePublicLeWM}{35/50}